\documentclass[11pt]{article}

\usepackage[utf8]{inputenc}
\usepackage[T1]{fontenc}
\usepackage{lmodern}
\usepackage{float}
\usepackage{placeins}
\usepackage{microtype}
\usepackage{amsmath,amssymb}
\usepackage{booktabs}
\usepackage{graphicx}
\usepackage{xcolor}
\usepackage[hidelinks]{hyperref}
\usepackage{tikz}
\usetikzlibrary{arrows.meta,positioning}

\usepackage[margin=1in]{geometry}

\title{\textbf{Same Story, Different Lens: A Multi-Agent System for Cross-Outlet News Framing Analysis}}

\author{%
  Affan Yasin\textsuperscript{1}, 
  Yueqi Lyu\textsuperscript{1}, 
  Rubia Fatima\textsuperscript{2},
  Weinchen Zhong\textsuperscript{1}, 
  Keyi Li\textsuperscript{2}\\[6pt]
  \parbox{0.96\textwidth}{\centering\small
    \textsuperscript{1} Academy of AI and Advanced Technology (AIAT),
    Xi’an Jiaotong-Liverpool University (XJTLU), Suzhou, China\\[3pt]
    \textsuperscript{2} Department of Computer Science and Technologies,
    Emerson University, Multan, Pakistan.}}

\date{}

\begin{document}

\maketitle

\begin{abstract}
News outlets covering the same event differ in what they emphasize and omit. \textit{Same Story, Different Lens} is an experimental research prototype for comparing these choices. It collects world and politics coverage from 17 outlets, groups reports about the same event, and produces structured
analyses of sentiment, political lean, framing, loaded language, possible omissions, and model-estimated proportions of factual claims and opinion. Loaded-language entries include quotations and neutral rewordings. Three model roles handle article analysis, assessment of source support, and cross-outlet briefing generation.
Analyses are stored in a JSONL archive. The evaluation compares outlet-level lean estimates with approximate, predefined reference ratings and summarizes the Critic's model-assigned grounding scores. These measures provide preliminary diagnostics. A grounding score reflects the model's assessment
of support in the supplied text and does not establish factual accuracy or constitute independent verification.
\end{abstract}

\vspace{0.5em}
\noindent\textbf{Keywords:} Multi-Agent Systems, News Framing Analysis, Large Language Models, Cross-Outlet News Analysis, Automated Source-Grounding Assessment.

\section{Introduction}
\label{sec:intro}

News framing concerns the selection and emphasis that shape a report of an event, as described by Entman~\cite{entman1993framing}. Comparing framing across outlets usually requires readers to examine several reports and identify differences in wording, emphasis, and missing context. Such comparisons are difficult to scale, particularly when readers need to trace an interpretation to the words that support it.

\textit{Same Story, Different Lens} groups reports about the same event and uses a language model to analyze each report. The prompt asks the model to support loaded-language judgments with quotations and framing judgments with specific wording. Readers can inspect these passages alongside the analysis.

The system provides:
\begin{enumerate}
  \item A working pipeline for news ingestion, story clustering, framing analysis, and presentation. A built-in mock provider supports end-to-end demonstrations without paid credentials.
  \item Separate Analyst, Critic, and Reporter roles. The Analyst generates structured annotations, the Critic assesses their support in the article, and the Reporter prepares a cross-outlet briefing.
  \item An archive and evaluation tools for examining outlet-level lean estimates and model-assigned grounding scores. The reproducibility design also specifies prompt tracking, schema versions, and field documentation.
\end{enumerate}

\begin{figure}[!htbp]
\begin{center}
\includegraphics[width=0.75\textwidth]{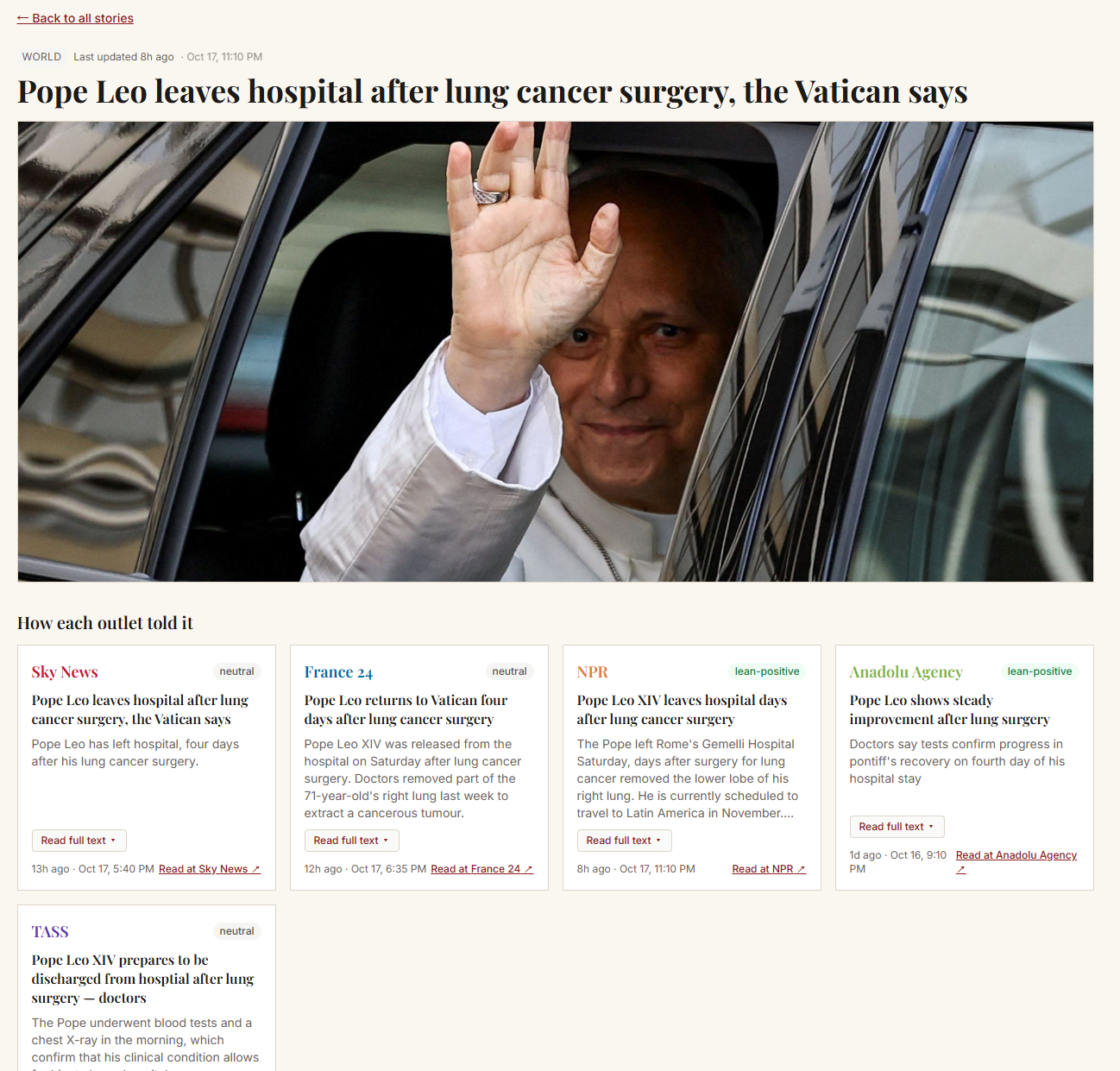}
\caption{Event-level comparison of reports about the same story.}
\label{figs:story}
\end{center}
\end{figure}

The reported evaluation covers 1{,}330 article analyses across 15 outlets. These results are preliminary. The reference ratings are approximate, and the model's annotations require review. Complete historical metadata are unavailable, which limits reconstruction of the reported evaluation. Figure~\ref{figs:story} shows the event-level comparison interface.

\section{Related work}
\label{sec:related}

\textbf{Framing and media slant.} Entman~\cite{entman1993framing} described framing in terms of selection and salience. Gentzkow and Shapiro~\cite{gentzkow2010media} found measurable slant in daily newspapers and an association with reader ideology. Hamborg et al.~\cite{hamborg2019automated} review automated methods that examine word choice, sentence structure, and topic selection. Media Bias/Fact
Check~\cite{mbfc} and AllSides~\cite{allsides} publish outlet-level media bias ratings.

\textbf{Comparing reports of the same event.} Framing comparisons require reports to be matched by event. Headline overlap and bag-of-words methods provide interpretable baselines for this task. The POLITICS approach~\cite{liu2022politics} uses pairs of articles about the same story to learn political leaning. Our system uses event matching to organize framing comparisons. It combines TF-IDF cosine similarity, headline Jaccard overlap, and union-find merging, without a neural text encoder.

\textbf{Language models as annotators and reviewers.} Language models can perform text annotation and, in some settings, outperform crowd workers~\cite{gilardi2023chatgpt}. They are also used to evaluate other models' outputs~\cite{zheng2023judging}. Their judgments have limitations: models can express political leanings~\cite{motoki2024human}, and greater model capability does not consistently prevent confident errors~\cite{zhou2024larger}.

Self-refinement methods ask a model to critique and revise its output~\cite{madaan2023selfrefine}. Constitutional approaches use explicit principles to constrain generation~\cite{bai2022constitutional}.
Verification methods assess support for generated statements; SelfCheckGPT, for example, uses sampling to detect possible hallucinations~\cite{manakul2023selfcheckgpt}. Our Critic assesses whether the supplied article supports the Analyst's annotations and records a model-assigned score and field-level comments. Its score is not a deterministically calculated proportion of verified claims.

The evaluation code uses established agreement measures: Cohen's $\kappa$~\cite{cohen1960coefficient}, adjusted Rand index~\cite{hubert1985comparing}, and Fowlkes--Mallows~\cite{fowlkes1983method}. The present snapshot lacks the human annotations needed to report these measures.
The contribution is a system that combines event matching, structured framing analysis, and inspectable review outputs. We do not claim a new clustering or statistical algorithm.

\begin{figure}[H]
  \centering
  \includegraphics[width=0.80\linewidth]{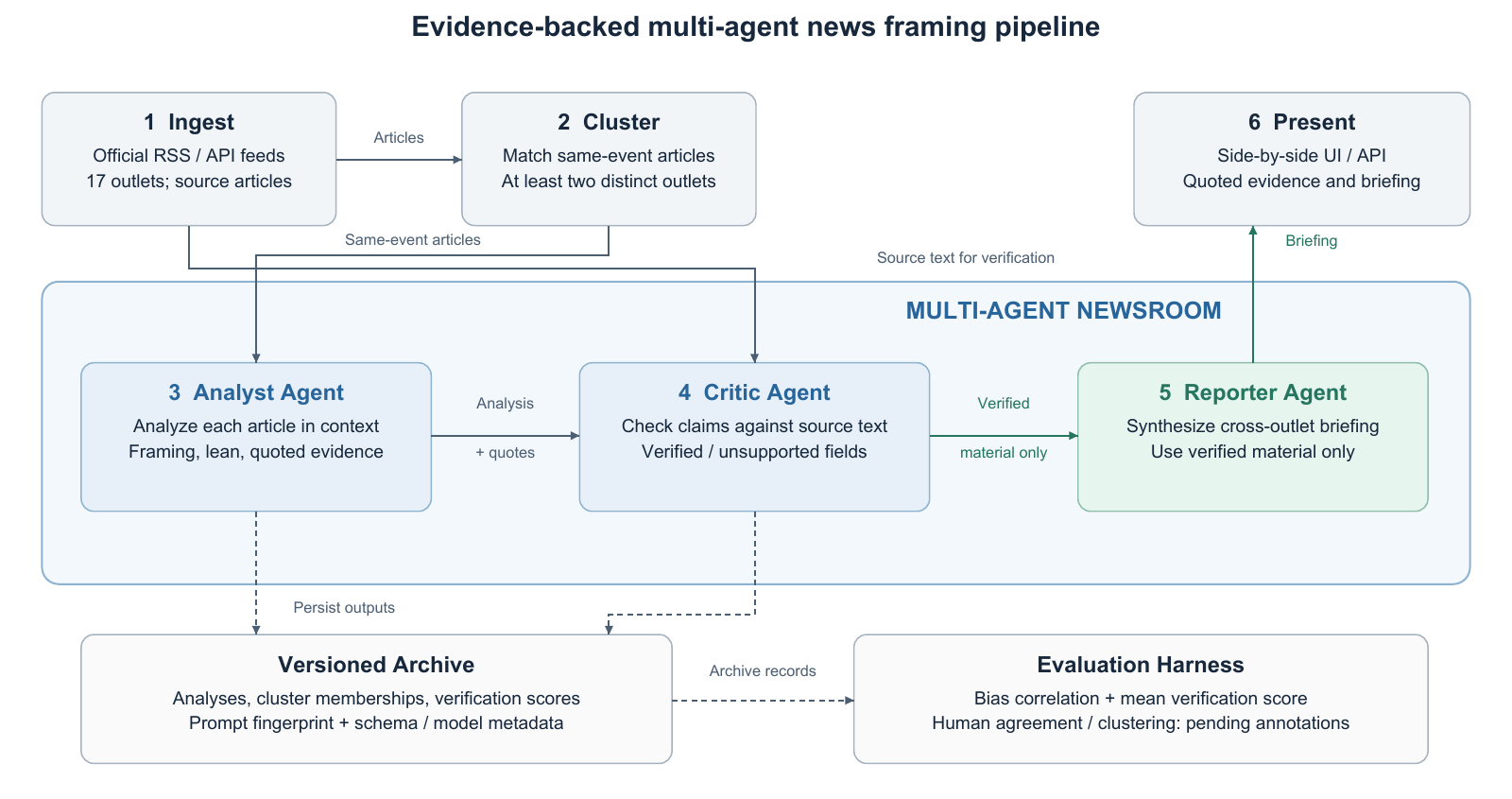}
  \caption{Design overview of the news-framing workflow. The Analyst generates annotations, the Critic assesses source support, and the Reporter prepares a briefing. The verified-input restriction and provenance mechanisms shown here are proposed design elements. The labels ``Verified material only'' and ``Use verified material only'' do not describe the current implementation, in which Critic assessments are diagnostic and do not filter Reporter inputs. Solid arrows show processing flows, and dashed arrows show archive and evaluation flows.}
  \label{fig:pipeline}
\end{figure}

\section{System design}
\label{sec:design}

The workflow covers ingestion, clustering, article analysis, assessment of source support, briefing generation, and presentation. Three roles organize the model tasks: the Analyst, Critic, and Reporter. Figure~\ref{fig:pipeline} shows the workflow and its archive and evaluation components.

An agent here is a task role. The reported evaluation uses a single underlying language model for these roles. The Analyst produces structured annotations, the Critic receives the article and those annotations, and the Reporter generates a briefing from story digests. This arrangement separates tasks but does not make their judgments independent.

The Analyst and Critic may share biases and failure modes. A field marked as supported records the Critic's judgment about the supplied text; it does not establish that the article itself is factually correct. In the current implementation, the Critic's assessments are stored as diagnostics and do
not filter the Reporter's inputs. The verified-input restriction shown in the design requires an explicit filtering mechanism.

The implementation is a TypeScript monorepo. An Express backend runs the pipeline and serves a JSON API, and a React frontend uses Vite and Tailwind. Clustering, evaluation metrics, and statistical calculations are written in TypeScript without heavy machine-learning dependencies.

\begin{figure}[h!]
\begin{center}
\includegraphics[width=0.80\textwidth]{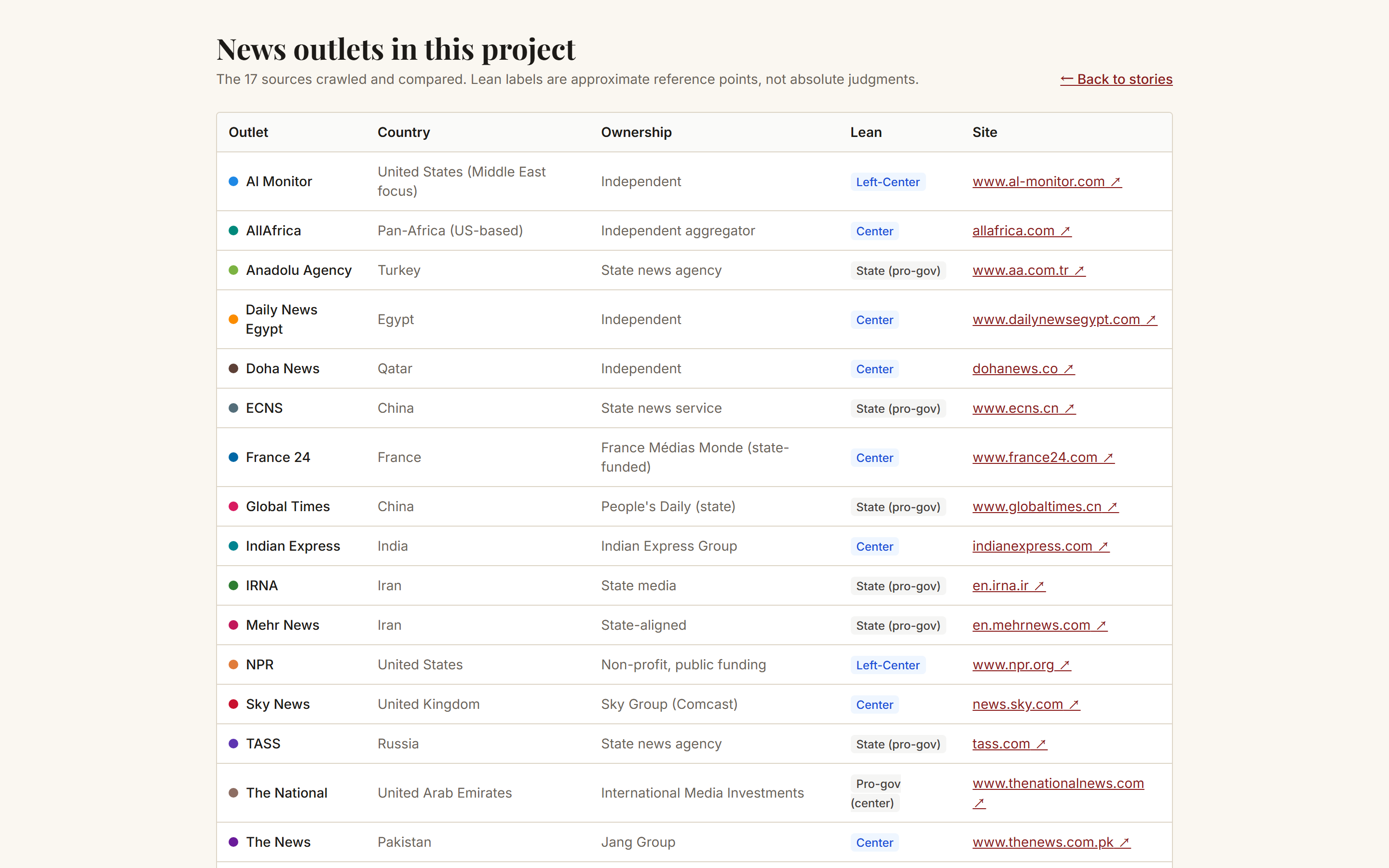}
\caption{The outlet directory and its display metadata. The screenshot's ``Lean'' column includes both political-lean descriptions and institutional relationships, which are distinct attributes. Directory labels
and the evaluation reference values in Table~\ref{tab:outlets}
should be distinguished.}

\label{figs:outlet}
\end{center}
\end{figure}

\subsection{Corpus and ingestion}
\label{sec:ingest}

The collection configuration covers 17 outlets through 19 official RSS feeds (Table~\ref{tab:outlets}). Sky News and NPR each contribute world and politics feeds; the other feeds cover world news. The selection includes public-service broadcasters such as NPR and France 24, commercial papers
such as Indian Express, news agencies such as Yonhap, and state-aligned media including TASS, Global Times, IRNA, Mehr News, Anadolu, and ECNS. Figure~\ref{figs:outlet} shows the outlet directory.

When an article body cannot be retrieved, the system can use the feed summary and record a \texttt{body\_incomplete} flag. Feed failures can also reduce the sources available for comparison. The requirement for at least two outlets in a story does not remove this source-selection risk.
Paywall and \texttt{robots.txt} compliance should be checked independently; the current ingestion implementation does not establish a general guarantee of compliance or reliable retention of the last successful crawl after every feed failure.

\FloatBarrier
\begin{table}[!htbp]
\centering
\small
\begin{tabular}{lll}
\toprule
\textbf{Outlet} & \textbf{Type} & \textbf{Reference lean}\\
\midrule
Sky News, France 24, The News, Doha News & broadcast / & center \\ 
The National, Daily News Egypt, AllAfrica, Yonhap & commercial & \\ \\
NPR, Al-Monitor, Indian Express & public / independent & lean-left \\ \\
Anadolu, ECNS & state agency & lean-right \\ \\
TASS, Global Times, IRNA, Mehr News & state media & right \\ \\
\bottomrule
\end{tabular}
\caption{The 17 configured outlets and their approximate, predefined reference ratings. These values are exploratory comparators, not independently validated ground truth; see Section~\ref{sec:eval}.}
\label{tab:outlets}
\end{table}

\begin{figure}[h!]
\begin{center}
\includegraphics[width=0.80\textwidth]{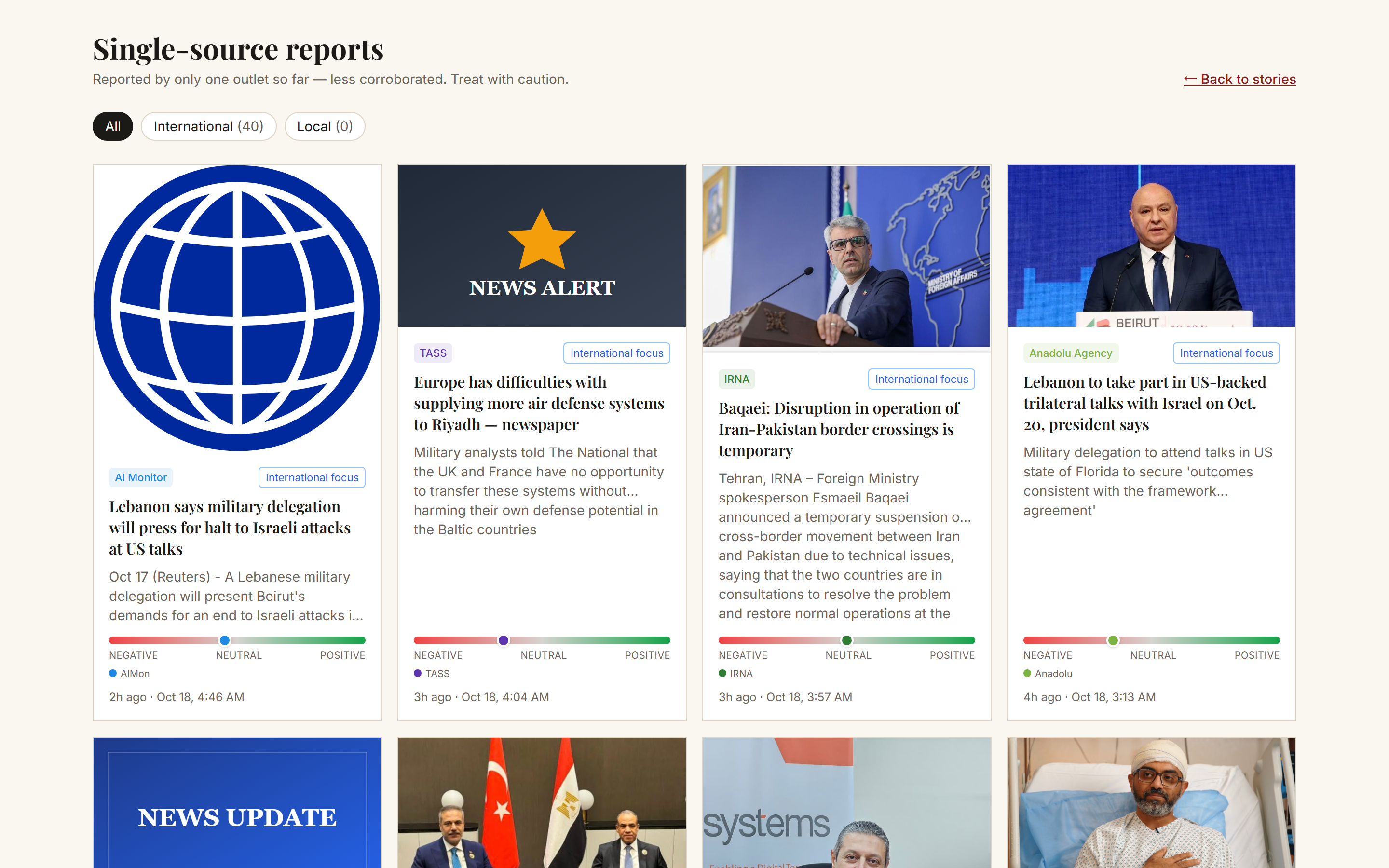}
\caption{Single-source reports retained separately when no second
outlet is matched.}
\label{figs:singles}
\end{center}
\end{figure}

\subsection{Story clustering}
\label{sec:cluster}

Each article has a TF-IDF representation of headline and summary unigrams and bigrams, with document frequencies calculated over the current crawl. Headline token sets provide a second similarity measure through Jaccard overlap. A pair of articles is linked when cosine similarity reaches $0.25$ or headline Jaccard similarity reaches $0.55$, provided the publication dates are no more than four days apart. Same-outlet pairs use higher thresholds of $0.45$ and $0.70$. Union-find merging converts these links into connected components. The four-day rule applies to article pairs, so a component formed through several links can span a longer interval.

A component becomes a story when it contains at least two distinct outlets. For a single-outlet component, the most recent report is retained in a separate list (Figure~\ref{figs:singles}). Stories are ranked first by the number of outlets, then by recency, and then by article count. The service retains the top 36 stories for presentation; the homepage initially displays 12. Analysis and archiving operate on the retained selection, so the archive does not contain every ingested article.

A fixed outlet-preference order selects the headline used as the topic label. This provides a consistent selection rule without establishing that the selected headline is neutral. Across crawls, URL and content
overlap link a story to an earlier instance so that follow-up reports can share an identifier. These links are heuristic and do not guarantee correct event identity.

\subsection{Per-article framing analysis (the Analyst)}
\label{sec:analyze}

The Analyst prompt asks for quotations to support analytical claims. When evidence is insufficient, the prompt instructs the model to leave the field empty and record an \texttt{insufficient\_\allowbreak evidence\_\allowbreak flags} entry. These are prompt requirements; they do not guarantee that every returned annotation has valid support. The output schema contains eight fields:

\begin{itemize}
  \item \texttt{sentiment}: a score in $[-1,1]$ and a five-way label from
        \texttt{negative} to \texttt{positive};
   \item \texttt{bias\_rating}: a model-assigned political-lean label from
        \texttt{left} to \texttt{right}, or \texttt{not-determinable},
        with a model-reported confidence value in $[0,1]$;
  \item \texttt{framing}: a one-sentence description of the article's angle;
    \item \texttt{loaded\_language}: up to five quotations, each with an explanation and a model-proposed neutral rewording;
  \item \texttt{omissions}: possible missing context identified by the model;
  \item \texttt{fact\_vs\_opinion}: model-estimated percentages of factual claims and opinion;
  \item \texttt{tone\_keywords}: up to five adjectives;
  \item \texttt{insufficient\_\allowbreak evidence\_\allowbreak flags}:
        fields for which support was insufficient.
\end{itemize}

The prompt defines bias in relation to other coverage and asks the model to describe wording and possible omissions without accusing an outlet of lying. However, the per-article Analyst does not receive the other outlets' full articles. Its lean and omission annotations therefore cannot be treated as direct measurements of differences established by reading all reports together.

System instructions and the JSON schema are separate from article text. This separation would allow the methodology to be fingerprinted independently of the input, as discussed in Section~\ref{sec:repro-infra}. A separate cluster-level comparison summarizes agreements and differences and identifies reports with relatively restrained or emotive wording.

The system reuses analyses through a URL-based cache and limits concurrent model requests. The cache does not establish that article content or method versions are unchanged. Individual analysis failures can produce deterministic fallback outputs labeled \texttt{mock}; these should be distinguished from real-model outputs in evaluation.

\subsection{Critic assessment and briefing synthesis}
\label{sec:newsroom}

The Critic receives the article text and the Analyst's output. It returns a \texttt{verification\_score} in $[0,1]$, lists of supported and unsupported fields, and suggested corrections. The model supplies the
score directly. The program does not derive it from a fixed count of supported fields, and the field lists do not determine its value. The Critic does not receive human annotations. We use its mean score as
an automated diagnostic of assessed source support.

The Reporter receives story digests containing topics, outlet counts, sentiment and lean summaries, claims shared across reports, and coverage differences. Its prompt requests a headline, a plain-language summary, one to four coverage divergences, and implications. The requested number of divergences is a prompt instruction rather than a strictly enforced output constraint. The current Reporter does not use the Critic's score or field lists to filter its inputs.

\begin{figure}[H]
\centering
\includegraphics[width=0.85\textwidth]{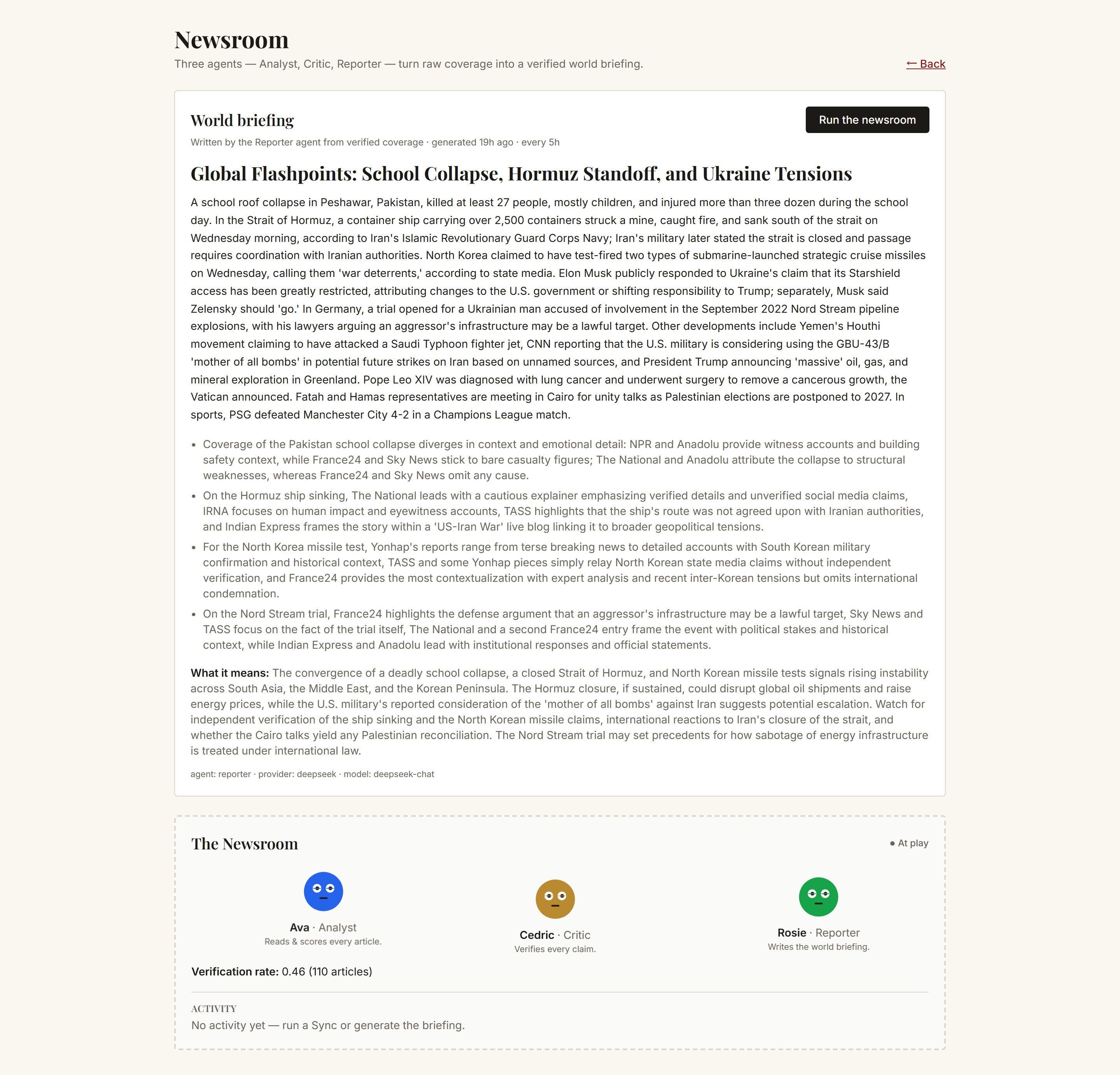}
\caption{The newsroom view, with a briefing and Critic statistics from the
live store. This illustrative screenshot does not represent the archived
evaluation subset.}
\label{figs:newsroom}
\end{figure}

Briefings run on a configurable schedule, with a default interval of five hours, rather than after every crawl. This limits repeated model requests but does not guarantee a successful briefing at every scheduled time. Figure~\ref{figs:newsroom} shows the newsroom view.

\subsection{Presentation layer}
\label{sec:present}
\begin{figure}[h!]
\begin{center}
\includegraphics[width=0.90\textwidth]{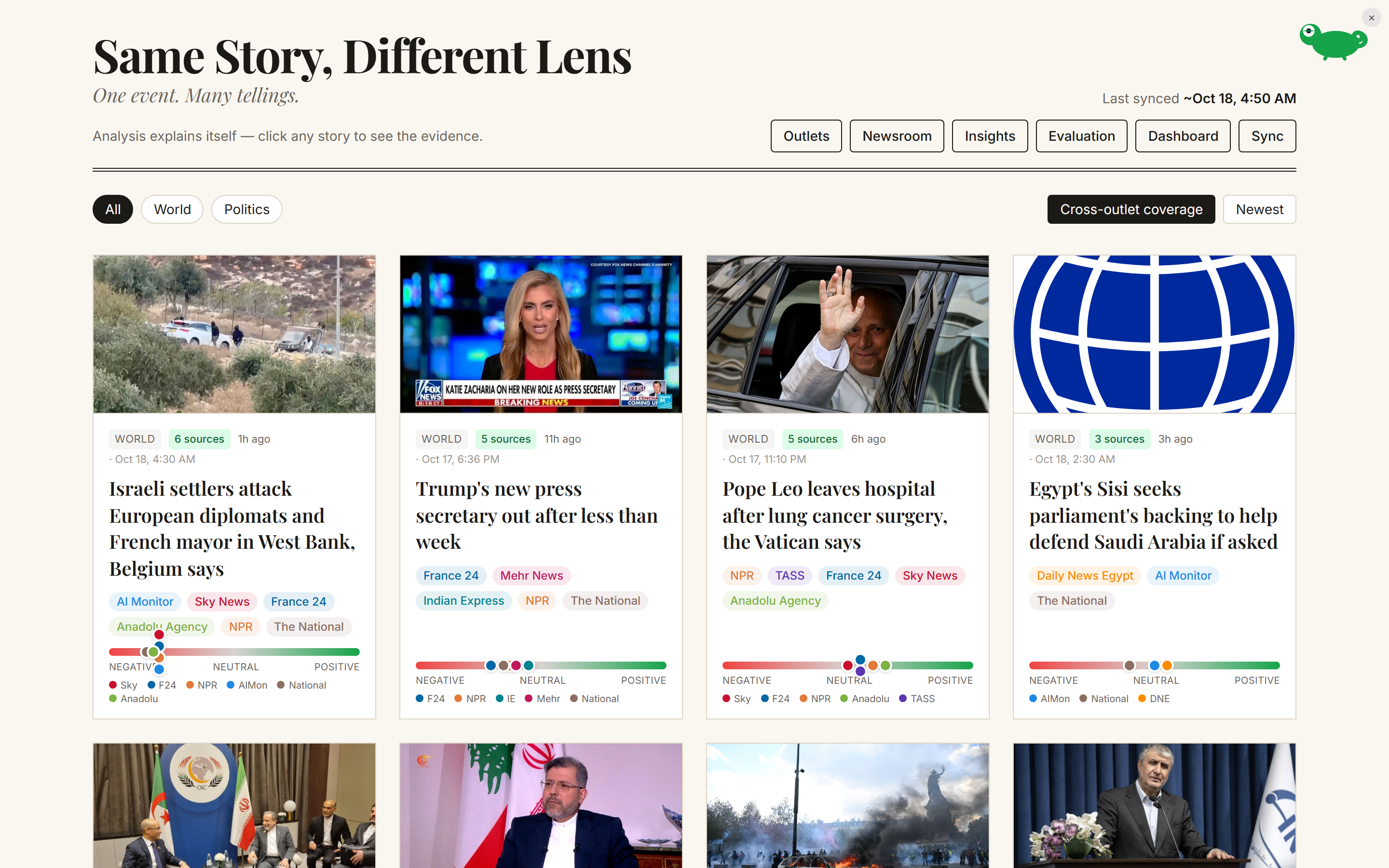}
\caption{The main feed ranks stories by outlet coverage and shows
sentiment summaries for each outlet.}
\label{figs:home}
\end{center}
\end{figure}

\begin{figure}[H]
\centering
\includegraphics[
  width=0.96\textwidth,
  height=0.90\textheight,
  keepaspectratio
]{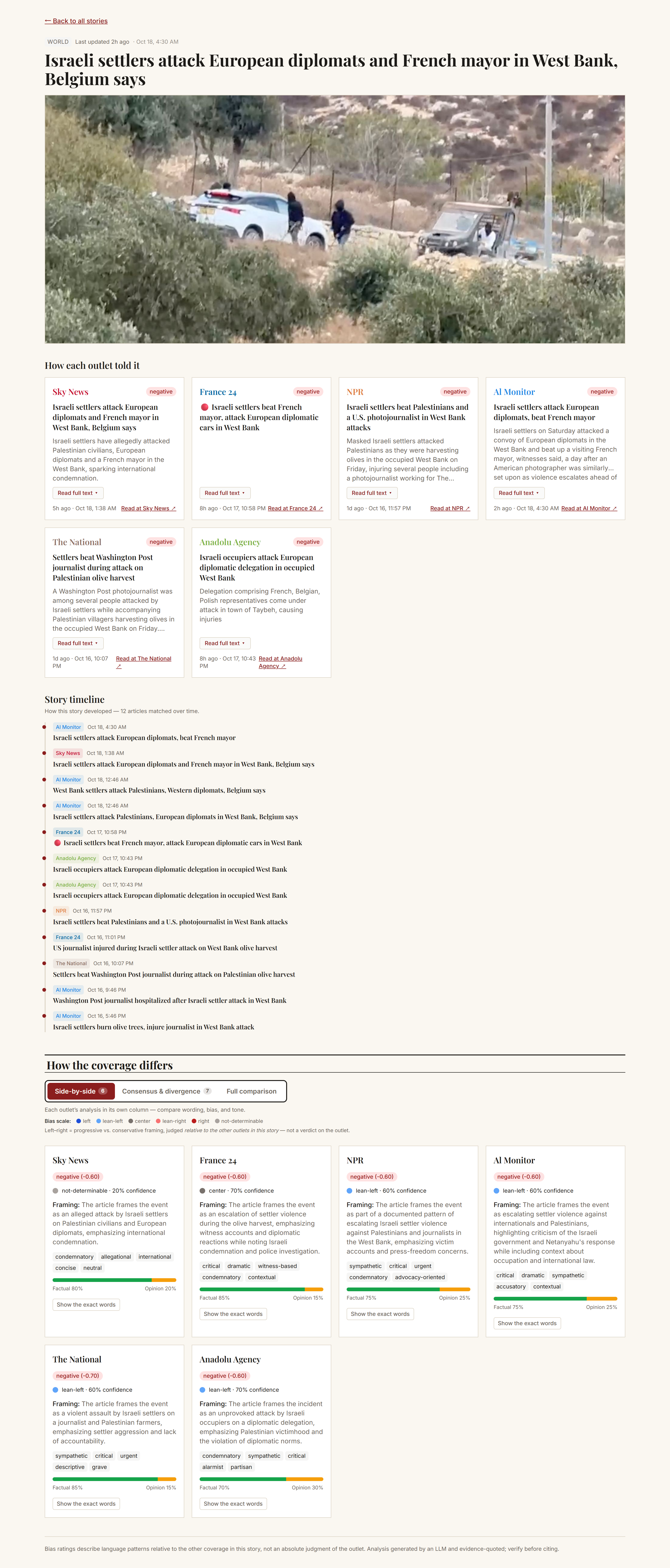}
\caption{The story-detail view for one cluster, with a divergence report
and an agreement matrix.}
\label{figs:sdf}
\end{figure}

The main feed lists stories with sentiment summaries for each outlet. A story-detail page places reports side by side and provides a divergence report, a model-generated ordering of reports from less to more emotive wording, and an agreement matrix.
Readers can expand loaded-language entries to inspect the quoted passages. Figures~\ref{figs:home} and Figure~\ref{figs:sdf} (Appendix A) show the feed and story-detail views.

The newsroom page presents briefings and live-store Critic statistics. A separate evaluation page displays archive-based measures. These views use different data scopes, so their displayed counts and averages need not match.

\FloatBarrier
\section{Reproducibility and evaluation}
\label{sec:repro}

\subsection{Archive and provenance}
\label{sec:repro-infra}

The JSONL archive stores analysis and verification outputs for later evaluation. The reproducibility design specifies three additional mechanisms:

\begin{enumerate}
  \item Prompt fingerprinting would identify system instructions and the schema with a SHA-256 hash and record the fingerprint and temperature for each analysis. Content-addressed snapshots and
        an index would preserve the corresponding prompts. 
        \item Schema versioning would record a semantic \texttt{schema\_version}, define how readers handle missing versions, and document changes in a changelog. The proposed default for a missing version is $1.0.0$.
  \item A codebook would document archive, store, and CSV fields, permitted labels, and flags.
\end{enumerate}

These mechanisms describe the provenance design. Their implementation and historical use are not established for the reported snapshot. Missing metadata cannot be reconstructed merely by assigning a version or hashing the current prompt. Full reconstruction of an earlier run requires its actual inputs, method version, and selection records.

\subsection{Evaluation harness}
\label{sec:eval}

The TypeScript evaluation code implements the following measures:

\begin{enumerate}
  \item Outlet-level lean correlation: means derived from the lean labels on $[-1,1]$ are compared with predefined reference ratings using Pearson $r$ and Spearman $\rho$.
  \item Inter-annotator agreement: Cohen's $\kappa$ for sentiment and lean when two sets of human annotations are available, together with model--human agreement.
  \item Topic-clustering agreement: adjusted Rand index and Fowlkes--Mallows between human \texttt{true\_topic} labels and cluster assignments.
  \item Mean Critic grounding score: the arithmetic mean of eligible \texttt{verification\_score} values.
\end{enumerate}

The reference ratings are approximate comparison values. Their presence in an evaluation file does not establish their provenance or independent validity, and it does not show that they were supplied to the Analyst or Critic for calibration.

A real-model evaluation requires explicit provider, model, duplicate, missing-label, and minimum-sample rules. The current default harness does not by itself enforce the reported exclusion of mock analyses or
the two-article threshold. It also rounds outlet means before computing correlations. Full-precision means should be used for a reproducible statistical calculation, with rounding applied only for display.

\begin{figure}[H]
\centering
\includegraphics[width=0.78\textwidth]{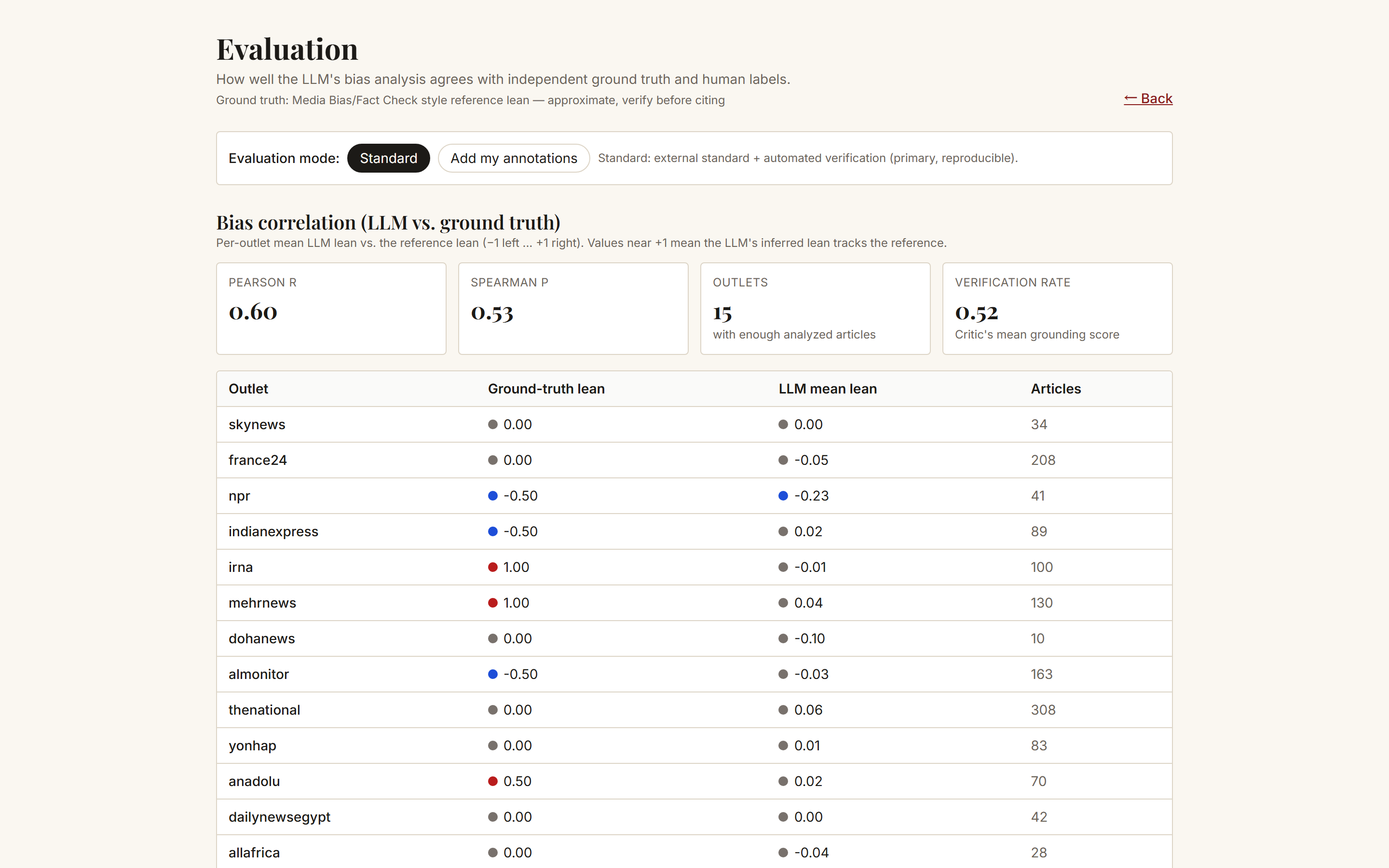}
\caption{The evaluation interface. This screenshot illustrates the view and does not reproduce the reported figures in Section~\ref{sec:results}. Human-agreement and clustering measures require annotations. The legacy labels ``ground truth'' and ``verification rate'' refer, respectively, to approximate reference ratings and the mean model-assigned Critic grounding score, not independently validated labels or a measured proportion of verified claims.}
\label{figs:eval}
\end{figure}

The treatment of \texttt{not-determinable} also needs to be stated. The implementation maps it to zero, which is not equivalent to a model judgment of \texttt{center}. Results can change if such labels
are excluded instead. The Critic mean likewise requires a declared scored subset and a separate report of scoring coverage. 

No human annotation file is available for the reported snapshot. Consequently, Cohen's $\kappa$ and the topic-clustering agreement measures remain pending. Their implementation still requires validation
before results from human annotations can be reported. Figure~\ref{figs:eval} shows the evaluation interface.

\section{Results}
\label{sec:results}

The original evaluation summary describes a snapshot accumulated over approximately two weeks. Its complete historical metadata and subset membership have not been recovered. We retain the reported figures below as preliminary results, rather than presenting them as an independent reproduction of the original experiment.

\begin{itemize}
  \item The collection configuration covers 17 outlets and 19 feeds. These are configured sources, not the number of sources necessarily included in every evaluation.
  \item The original summary reports 3{,}174 archived article records and 1{,}330 DeepSeek analysis records. Record counts should not be interpreted as counts of unique articles or model calls.
        Mock and unknown-source outputs require separate accounting; the remaining records cannot all be assumed to be mock. 
    \item The reported lean-correlation subset contains 15 outlets, each described as having at least two real-model analyses. Its exact selection has not been reconstructed.
  \item The reported Critic subset contains 856 scored real-model analyses. The score mean refers to these records, not all 1{,}330 analyses. The reasons for missing scores have not been established.
\end{itemize}

\textbf{Lean correlation.} Table~\ref{tab:lean} reproduces the reported outlet means and reference ratings. The original summary gives Pearson $r = 0.76$ and Spearman $\rho = 0.65$ across 15 outlets. TASS and Mehr News have positive model means, while Indian Express is close to the center despite its lean-left reference rating. These comparisons describe the displayed values and do not validate the underlying reference ratings.

\begin{figure}[h!]
  \centering
  \includegraphics[width=0.78\textwidth]{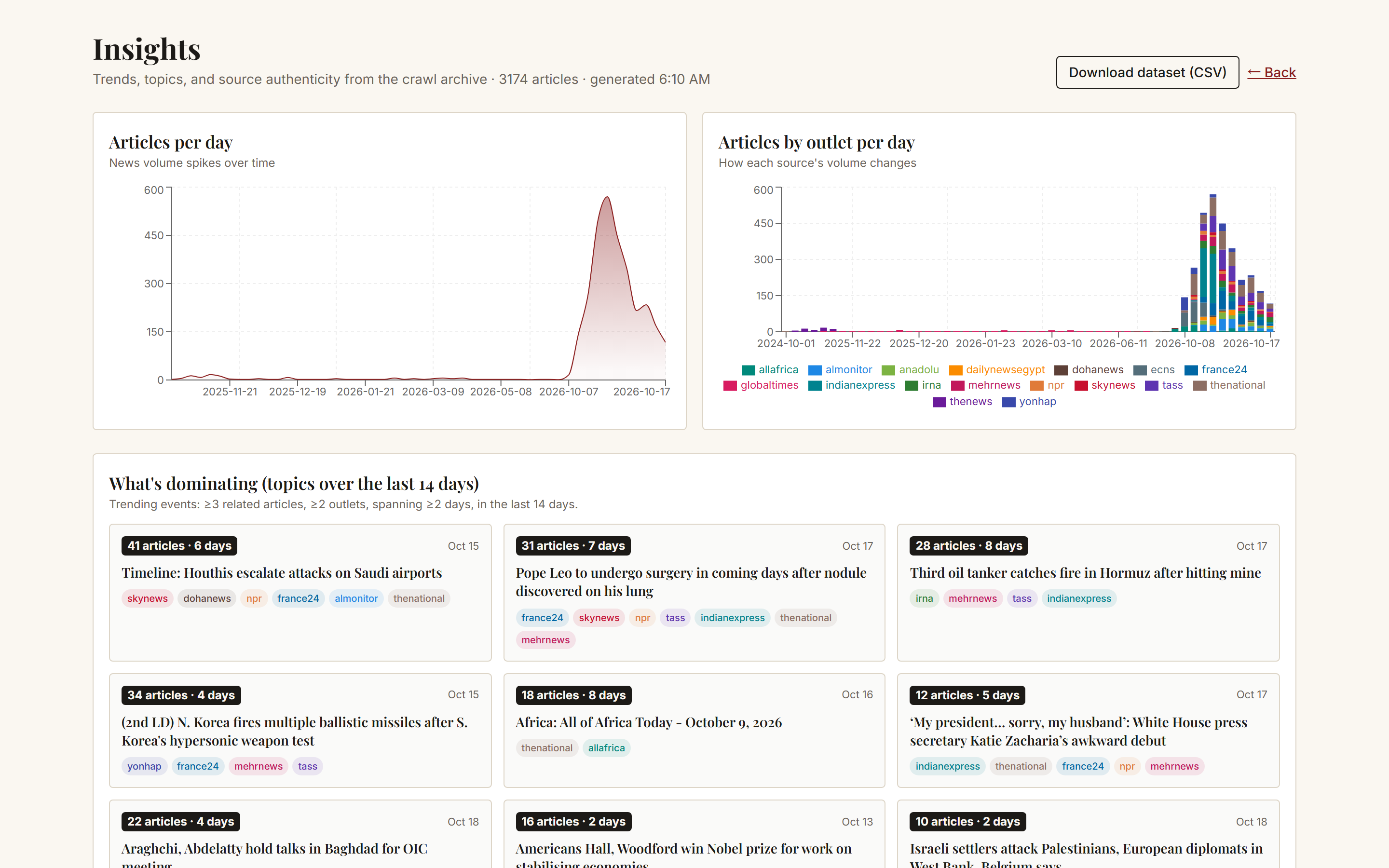}
  \caption{The Insights view, showing article-volume trends,
  coverage by outlet, and trending topics.}
  \label{figs:insights}
\end{figure}

The displayed article counts sum to 697, leaving 633 records unaccounted for relative to the reported 1{,}330 analyses. Without the original selection records, this difference cannot be assigned to a particular filter or deduplication rule. Full-precision inputs are also needed to reconstruct the reported correlations; the rounded table alone does not establish their original calculation.

\begin{table}[h!]
\centering
\small
\begin{tabular}{lrrr}
\toprule
\textbf{Outlet} & \textbf{Reference lean} & \textbf{Model lean} & \textbf{$n$}\\
\midrule
Al-Monitor & $-0.50$ & $-0.03$ & 123\\
Anadolu & $+0.50$ & $+0.00$ & 34\\
Daily News Egypt & $0.00$ & $+0.03$ & 18\\
ECNS & $+0.50$ & $+0.10$ & 10\\
France 24 & $0.00$ & $-0.07$ & 81\\
Indian Express & $-0.50$ & $+0.03$ & 34\\
IRNA & $+1.00$ & $+0.25$ & 2\\
Mehr News & $+1.00$ & $+0.29$ & 29\\
NPR & $-0.50$ & $-0.28$ & 23\\
Sky News & $0.00$ & $-0.33$ & 3\\
TASS & $+1.00$ & $+0.50$ & 98\\
The National & $0.00$ & $+0.06$ & 172\\
Yonhap & $0.00$ & $+0.00$ & 55\\
AllAfrica & $0.00$ & $-0.09$ & 11\\
Doha News & $0.00$ & $-0.13$ & 4\\
\midrule

\multicolumn{4}{l}{Reported correlations: Pearson $r = 0.76$, Spearman $\rho = 0.65$; 15 outlets}\\
\bottomrule
\end{tabular}
\caption{Reported outlet-level model means and reference ratings on $[-1,1]$, with negative values indicating left lean. $n$ is the reported number of included analyses. Each outlet contributes one observation to the unweighted correlation; article counts are not correlation weights. Means are displayed to two decimal places. The full historical selection and precision are unavailable.}
\label{tab:lean}
\end{table}

\textbf{Critic grounding score.} The reported mean over 856 scored analyses is $0.52$. This is an average of model-assigned scores. It does not mean that half of all analytical claims were independently
verified. Supported and unsupported field lists can help identify outputs for review, but the historical field-level counts are insufficient to support a general ranking of which fields were most reliably grounded.

\textbf{Human agreement.} No annotation file accompanies the reported snapshot. Cohen's $\kappa$ and topic-clustering agreement therefore remain pending.

\begin{figure}[h!]
  \centering
  \includegraphics[width=0.78\textwidth]{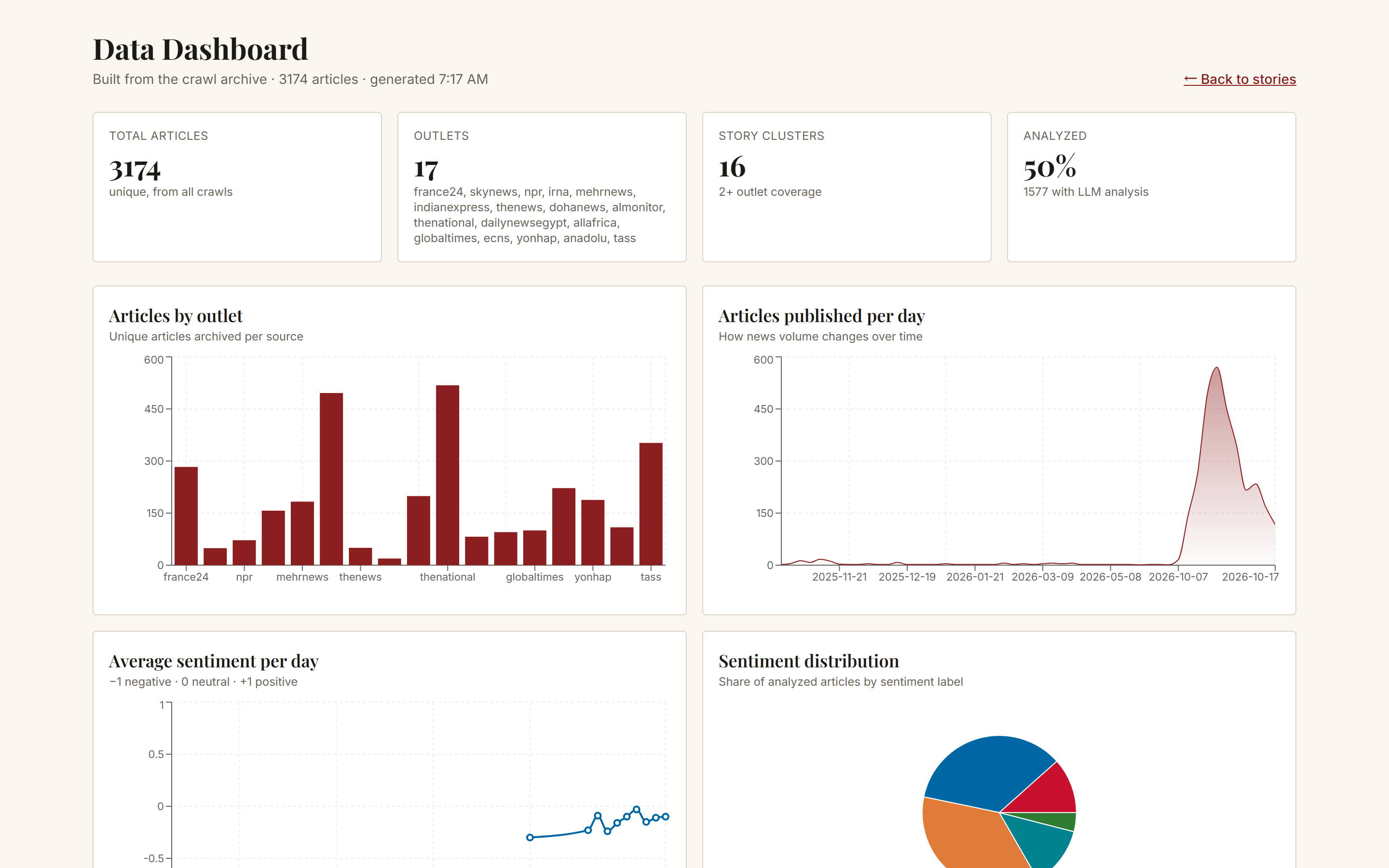}
  \caption{The dashboard, showing archive counts, outlet coverage,
  current story-cluster counts, and sentiment summaries. The analyzed
  count includes records with an analysis object, including mock outputs.}
  \label{figs:dashboard}
\end{figure}

The limited observation window and approximate reference ratings restrict interpretation. The results do not establish an outlet ranking or the accuracy of individual article labels. Figures~\ref{figs:insights} and~\ref{figs:dashboard} illustrate the archive views; their screenshots are not additional evidence for the reported correlation and grounding scores.

\section{Limitations and ethics}
\label{sec:limits}
\textbf{Limitations.} The reported evaluation covers fifteen outlets, several represented by only a few articles, and a single underlying model. The reference ratings are approximate and predefined. Their sources and assignment procedures require further verification. Correlations describe association and relative ordering, not absolute agreement in lean scores. The Analyst and Critic may share systematic
biases and errors. A Critic score records the model's assessment of support in the supplied text; it does not establish factual correctness or independent verification. Incomplete historical metadata limits
reconstruction of earlier evaluations, and repeated records and unspecified selection rules can affect the reported means. Lightweight lexical clustering can separate reports of the same event when their wording differs or merge distinct events when their wording is similar.

\textbf{Ethics.} The interface presents quoted passages and possible omissions so that readers can examine the basis of an interpretation. It treats lean as relative to other coverage and identifies the
reference ratings as approximate. Such tools can still be used to make unsupported accusations of partisanship. Quotation access and instructions against accusations of lying are intended to support
review; they do not ensure that every output is defensible. Related work suggests that carefully designed AI interactions can reduce unfounded beliefs~\cite{costello2024durably}, but that finding
does not establish the effect of this prototype.

The archive is released under a non-commercial research license (CC-BY-NC for the data, with the code under a corresponding non-commercial license).

\textbf{Prototype status.} Summaries and framing analyses are generated automatically, and aggregate statistics are computed programmatically. The original articles come from external publishers. Outputs may contain errors, omissions, or biased interpretations and should be checked against the linked sources. This is an experimental research prototype, with additional functionality under development. Future work will extend the existing evaluation harness through stronger provenance tracking, broader testing, and independent human assessment.

\section{Conclusion}
\label{sec:conclusion}

\textit{Same Story, Different Lens} brings event clustering, structured framing annotations, Critic diagnostics, and cross-outlet briefings into one prototype. Readers can inspect quotations and intermediate outputs alongside the generated analysis.

The reported outlet correlation ($r = 0.76$) and mean Critic score ($0.52$) are preliminary. The original evaluation requires further work on snapshot identity, subset selection, and reference-rating
provenance. The Critic's score measures model-assessed source support, and the current workflow does not restrict Reporter inputs to Critic-approved material. Further evaluation needs independent human
annotations and a fully traceable archive.

\section*{Availability}
\label{sec:availability}
The source code and data schema are available at Github \url{(https://github.com/affanyasin1/multi-perspective-analyzer)}\footnote{https://github.com/affanyasin1/multi-perspective-analyzer.}.
\vspace{2mm}
\section*{Acknowledgments.} 
This work benefited from AI-assisted tools for language editing, improving the clarity and readability of the manuscript, and supporting code development.
\bibliographystyle{plain}

\section*{Appendix A}

\end{document}